\pdfoutput=1

\documentclass[11pt]{article}

\usepackage{EMNLP2023}

\usepackage{times}
\usepackage{latexsym}

\usepackage[T1]{fontenc}

\usepackage[utf8]{inputenc}

\usepackage{microtype}

\usepackage{inconsolata}

\usepackage{booktabs}
\usepackage{graphicx} 
\usepackage{multirow}
\usepackage{color}
\newcommand{\Sec}[1]{{Section~\ref{#1}}}
\newcommand{\Tab}[1]{{Table~\ref{#1}}}

\makeatletter
\newcommand{\reffnmark}[1]{%
    \begingroup
        \unrestored@protected@xdef\@thefnmark{\ref{#1}}%
    \endgroup
    \@footnotemark}
\makeatother
\newcommand{\unidir}[2]{{#1$\rightarrow$#2}}

\title{Unsupervised Translation Quality Estimation \\ Exploiting Synthetic Data and Pre-trained Multilingual Encoder}

\author{
    Yuto Kuroda${}^{\dag}$ \quad Atsushi Fujita${}^{\ddag}$ \quad Tomoyuki Kajiwara${}^{\dag}$ \quad Takashi Ninomiya${}^{\dag}$ \\
    ${}^{\dag}$Ehime University \\
    ${}^{\ddag}$National Institute of Information and Communications Technology \\
    \texttt{\{kuroda@ai., kajiwara@, ninomiya@\}cs.ehime-u.ac.jp} \\
    \texttt{atsushi.fujita@nict.go.jp} \\
    }

\begin{document}
\maketitle
\begin{abstract}
  Translation quality estimation (TQE) is the task of predicting translation quality without reference translations.
  Due to the enormous cost of creating training data for TQE, only a few translation directions can benefit from supervised training.
  To address this issue, unsupervised TQE methods have been studied.
  In this paper, we extensively investigate the usefulness of synthetic TQE data and pre-trained multilingual encoders in unsupervised sentence-level TQE, both of which have been proven effective in the supervised training scenarios.
  Our experiment on WMT20 and WMT21 datasets revealed that this approach can outperform other unsupervised TQE methods on high- and low-resource translation directions in predicting post-editing effort and human evaluation score, and some zero-resource translation directions in predicting post-editing effort.
\end{abstract}

\section{Introduction}

Translation quality estimation (TQE) \citep{blatz-etal-2004-confidence,specia-2018} is the task to predict translation quality of an output of machine translation (MT) system with respect to the source sentence.
Developing TQE methods that correlate well with human evaluation can support human decision-making: whether to use the MT output as is, to post-edit it, or to use other MT systems.
Hence, TQE is practically important to advance the real-world use of MT systems.

Methods for TQE have been actively studied mainly in the WMT TQE shared task \cite{specia-etal-2020-findings-wmt,specia-etal-2021-findings, zerva-etal-2022-findings}.
Most of them are supervised methods, i.e., the TQE models are trained on tuples of source sentence, MT output, and manually labeled quality score.
However, creating a TQE dataset, i.e., a set of such tuples, is costly because the quality score can only be obtained through manual work, such as human evaluation for DA (direct assessment) score or post-editing for HTER (Human-targeted Translation Edit Rate) \citep{snover-etal-2006-study}, carried out by annotators who have high competence in both source and target languages.
Furthermore, the distribution of translation quality and errors can vary widely from one MT system to another; thus it is impractical to tailor such a dataset for each MT system.
As such, even in the latest WMT22 TQE task \citep{zerva-etal-2022-findings}, human-labeled training data are available only for eleven translation directions.

To address this issue, unsupervised TQE methods have been studied.
As reviewed in \Sec{sec:previous}, past work has exploited pre-trained multilingual encoders and neural MT (NMT) systems, as well as bilingual parallel data.
Synthetic TQE data generated from monolingual or parallel data have also been exploited in order to adapt given pre-trained encoders to the TQE task \citep{tuan-etal-2021-quality,eo-etal-2021-dealing}.
However, no study has yet made an extensive investigation into the impact of the synthetic TQE data on relatively recent pre-trained multilingual encoders that are prevalent in the supervised training scenarios \citep{ranasinghe-etal-2020-transquest,wang-etal-2021-qemind,rei-etal-2022-cometkiwi}.

This paper investigates the usefulness of synthetic TQE data in unsupervised training of sentence-level TQE models on top of relatively recent pre-trained multilingual encoders.
Our experiment on WMT20 and WMT21 datasets shows that this approach can outperform other unsupervised TQE methods on high- and low-resource language pairs.
Even though each component method in this study is not novel, it makes the first comparison of representative options for unsupervised TQE.

\section{Previous Work}
\label{sec:previous}

In this paper, we regard TQE methods that do not use any human-labeled pairs of source sentence and an MT output for it as ``unsupervised.''

Existing unsupervised TQE methods rely on either multilingual encoders or NMT models.
Unsupervised methods based on a multilingual encoders compute cosine similarity between sentence embeddings of source sentence and MT output as the TQE score \cite{fonseca-etal-2019-findings, song-etal-2021-sentsim}.
For example, TQE using only LaBSE \citep{feng-etal-2022-language} can be considered as an unsupervised method,
since LaBSE has been trained through fine-tuning mBERT \citep{devlin-etal-2019-bert} on parallel data, where mBERT is pre-trained exclusively on monolingual data in multiple languages.
Sentence embeddings obtained by LaBSE are dominated by language-specific information, and thus hinder precise estimation of semantic similarity across different languages \citep{tiyajamorn-etal-2021-language}.
DREAM \citep{tiyajamorn-etal-2021-language} and MEAT \citep{kuroda-etal-2022-adversarial} address this issue by disentangling sentence embeddings to meaning and language embeddings, and outperformed LaBSE in the sentence-level TQE task by comparing meaning embeddings.

In addition to parallel data, synthetic TQE data have also been exploited for TQE.
\citet{tuan-etal-2021-quality} generated synthetic TQE data from a parallel corpus using either an MT system or a masked language model.
\citet{eo-etal-2021-dealing} did so from monolingual data in the target language via round-trip MT.
However, existing work has not assessed the impact of synthetic TQE data on training models with multilingual encoders; \citet{tuan-etal-2021-quality} evaluated only the predictor-estimator model based on bidirectional LSTM \citep{kim-etal-2017-predictor}.
\citet{eo-etal-2021-dealing} trained TransQuest with a relatively new multilingual encoder, i.e., \mbox{XLM-R$_{\mathrm{Large}}$} \citep{conneau-etal-2020-unsupervised}, but did not evaluate their method on human-labeled data.
Given the promising results in the supervised TQE scenarios obtained with more sophisticated pre-trained multilingual encoders \citep{ranasinghe-etal-2020-transquest, specia-etal-2020-findings-wmt, specia-etal-2021-findings, wang-etal-2021-qemind, zerva-etal-2022-findings, rei-etal-2022-cometkiwi}, it is worth investigating whether such models can also be trained well in a fully unsupervised fashion, i.e., without using any human-labeled TQE data.

Other unsupervised TQE methods rely on NMT models trained on parallel data.
While \citet{fomicheva-etal-2020-unsupervised} used internal weights of NMT models, Prism \citep{thompson-post-2020-automatic} used only the forced-decoding score, i.e., the likelihood of MT output given source sentence computed by the given NMT model, as the TQE score.
Publicly available pre-trained multilingual NMT models, such as mBART50 \citep{tang-etal-2021-multilingual} and M2M-100 \citep{fan-etal-2022-beyond}, extend the applicability of this approach to many translation directions.

\section{Training with Synthetic TQE Data}
\label{sec:methods}

In this paper, we primarily focus on HTER \cite{snover-etal-2006-study} as the sentence-level quality metric.
HTER is the edit rate of MT outputs with respect to their human post-edited versions.
HTER has drawn less attention \citep{tuan-etal-2021-quality,eo-etal-2021-dealing}; most of past work targeted DA scores \citep{fomicheva-etal-2020-unsupervised, thompson-post-2020-automatic, tiyajamorn-etal-2021-language, kuroda-etal-2022-adversarial}.  Nevertheless, HTER is also a useful indicator in the translation production workflow where MT output is regarded as a draft translation \citep{ISO18587:17} and thus has been tackled in WMT \cite{specia-etal-2020-findings-wmt,specia-etal-2021-findings, zerva-etal-2022-findings}.
We also assume that we can obtain synthetic TQE data that approximate it as explained below.  In contrast, there is no such a way for DA scores.

\paragraph{Generation of synthetic TQE data:}

Following \citet{tuan-etal-2021-quality}, we generate synthetic TQE data from a bilingual parallel corpus, $C=(x_{i},y_{i})_{i=1}^{N}$.
First, using an MT system, we translate the source sentences of the parallel corpus, $x_{i}$, into the target language.
We then compute translation edit rate (TER) \cite{snover-etal-2006-study} for each MT output, $y'_{i}$, and its corresponding human translation in the parallel corpus, $y_{i}$.
Finally, we compose synthetic TQE data, comprising $(x_{i},y'_{i},\mathrm{TER}(y_{i},y'_{i}))_{i=1}^{n}$.
Here, we assume that TER would approximate HTER even though $y$ and $y'$ are separately generated.

\paragraph{TQE model:}

We investigate neural TQE models consisting of a pre-trained multilingual encoder and additional layers.
In particular, we compare the following two encoding methods, following \citet{ranasinghe-etal-2020-transquest}.
\begin{description}\itemsep=0mm
  \item[Split:] Source sentence and MT output are separately encoded with a prepended [CLS] token.
    In other words, we encode ``[CLS] source sentence'' and ``[CLS] MT output'' respectively, concatenate the two embeddings each corresponding to the [CLS] token, and feed the following layers with it.
  \item[Concat:] Source sentence and MT output are jointly encoded with a prepended [CLS] token and a [SEP] token in between.
    That is, we encode only ``[CLS] source sentence [SEP] MT output'' and input the embedding corresponding to the [CLS] token to the following layers.
\end{description}
In both methods, we input the embedding(s) corresponding to the [CLS] token(s) to the following multi-layer perceptron (MLP),
and train a regression model through minimizing root mean squared error (RMSE).

\section{Experiment}
\label{sec:experiment}

We evaluated the usefulness of synthetic TQE data with the methods presented in \Sec{sec:methods} on the WMT20 and WMT21 TQE datasets \citep{specia-etal-2020-findings-wmt, specia-etal-2021-findings, fomicheva-etal-2022-mlqe}.
As the metrics to predict, we used both HTER and DA score.

\subsection{Setting}

\paragraph{Test sets for HTER:}

The WMT20 TQE Task 2 dataset\footnote{\url{https://statmt.org/wmt20/quality-estimation-task.html}\label{fn:WMT20-QE}} includes two high-resource translation directions: English-to-German (\unidir{En}{De}) and English-to-Chinese (\unidir{En}{Zh}).
Instead of official test sets for which gold HTER scores were not disclosed, we used the official development sets consisting of $1$k tuples for our evaluation.
For the WMT21 TQE Task 2 dataset,\footnote{\url{https://statmt.org/wmt21/quality-estimation-task.html}\label{fn:WMT21-QE}} we used all the test sets for 11 translation directions, each consisting of $1$k tuples of source sentence, an MT output for it, and HTER score obtained through post-editing the MT output.
Note that lower HTER indicates higher quality of MT output.

\paragraph{Test sets for DA score:}

We used the WMT20 TQE Task 1\reffnmark{fn:WMT20-QE} and WMT21 TQE Task 1\reffnmark{fn:WMT21-QE} datasets.
The former contains $1$k tuples of source sentece, an MT output for it, and a DA score for each of six translation directions,
while the latter consists of another sets of $1$k tuples for the same six translation directions and five translation directions for which no parallel data are available for TQE.
The DA score has been computed by standardizing 0--100 scores given by multiple annotators.
Note that higher DA score indicates higher MT quality.

\paragraph{Classification of translation directions:}

The MT outputs for each translation direction had been generated by a Transformer model \citep{vaswani-2017} trained using the fairseq toolkit\footnote{\url{https://github.com/pytorch/fairseq}} \citep{ott-2019}.\footnote{\url{https://github.com/facebookresearch/mlqe/tree/main/nmt_models}}
Henceforth, according to the size of bilingual parallel corpora used to train the MT models
(\Tab{table:num-data}) and we exploit for synthesizing TQE data, we denote
\unidir{En}{De} and \unidir{En}{Zh} as high-resource, Romanian-to-English (\unidir{Ro}{En}) and Estonian-to-English (\unidir{Et}{En}) as medium-resource, Nepali-to-English (\unidir{Ne}{En}) and Sinhalese-to-English (\unidir{Si}{En}) as low-resource, and the rest as zero-shot\footnote{Note that bilingual parallel data for these language pairs could have been used for pre-training.} translation directions.

\begin{table}[t]
  \centering
  \small
  \begin{tabular}{cr@{}c@{}lrr}
    \toprule
    \multicolumn{4}{c}{Language pair} & \multicolumn{1}{c}{Bilingual} & \multicolumn{1}{c}{Synthetic}                                \\
    \midrule
                                      & En                            & --                            & De & 23,360,441 & 22,268,661 \\
                                      & En                            & --                            & Zh & 20,305,268 & 19,811,665 \\
    \midrule
                                      & Ro                            & --                            & En & 3,901,501  & 3,178,631  \\
                                      & Et                            & --                            & En & 877,769    & 863,973    \\
    \midrule
                                      & Ne                            & --                            & En & 498,271    & 477,563    \\
                                      & Si                            & --                            & En & 646,766    & 585,598    \\
    \bottomrule
  \end{tabular}
  \caption{The number of sentence pairs in the bilingual parallel corpora and synthetic TQE data.}
  \label{table:num-data}
\end{table}
%

\paragraph{Synthetic TQE data:}

To generate synthetic TQE data, we used M2M-100\footnote{\url{https://huggingface.co/facebook/m2m100_418M}} \citep{fan-etal-2022-beyond} and the bilingual parallel corpora officially provided by the task organizers\reffnmark{fn:WMT21-QE} (\Tab{table:num-data}) on which the aforementioned MT systems had been trained.
Before decoding the source-side of the parallel corpora, we fine-tuned M2M-100 for each translation direction.
First, we sampled $1$M, $200$k, and $50$k sentence pairs for high-, medium-, and low-resource language pairs, respectively, from the bilingual parallel corpora (\Tab{table:num-data}),
where we limited parallel sentences to those whose each side was composed of up to $128$ sub-word tokens.
We then fine-tuned M2M-100 on the sampled parallel sentences with the AdamW optimizer \citep{loshchilov2018decoupled} ($\beta_{1}=0.9$, $\beta_{2}=0.999$, $\epsilon=1\times 10^{-8}$) on batches of $16$ sentence pairs with a learning rate of $3\times 10^{-5}$, and terminated training after one epoch for each of high- and medium-resource language pairs and three epochs for each of low-resource language pairs.
When decoding the source-side of the bilingual parallel corpora, we performed beam search with beam size of 5, length penalty of 1.0, and maximum sequence length of 200.
After decoding, we calculated TER of the MT output with respect to the target-side of the parallel corpus using SacreBLEU\footnote{\url{https://github.com/mjpost/sacrebleu}} \citep{post-2018-call}, and composed synthetic tuples of source sentence, MT output, and the TER score, discarding those with TER score greater than $1.0$.
See \Tab{table:num-data} for the results.

\begin{table}[t]
  \centering
  \small
  \begin{tabular}{lcc}
    \toprule
    Model                                           & $P$  & $A$   \\
    \midrule
    LaBSE                                           & 470M & -     \\
    MEAT                                            & 470M & 1M    \\
    M2M-100                                         & 490M & -     \\
    \midrule
    XLM-R$_{\mathrm{Base}}$+MLP (split)             & 280M & 1,537 \\
    XLM-R$_{\mathrm{Large}}$+MLP (split)            & 560M & 2,049 \\
    \textsc{InfoXLM}$_{\mathrm{Base}}$+MLP (split)  & 280M & 1,537 \\
    LaBSE+MLP (split)                               & 470M & 1,537 \\
    \midrule
    XLM-R$_{\mathrm{Base}}$+MLP (concat)            & 280M & 769   \\
    XLM-R$_{\mathrm{Large}}$+MLP (concat)           & 560M & 1,025 \\
    \textsc{InfoXLM}$_{\mathrm{Base}}$+MLP (concat) & 280M & 769   \\
    LaBSE+MLP (concat)                              & 470M & 769   \\
    \bottomrule
  \end{tabular}
  \caption{Number of parameters: $P$ for the pre-trained model and $A$ for the additional MLP component(s).}
  \label{tab:model_param}
\end{table}
\begin{table*}[t]
  \centering
  \small
  \begin{tabular}{lccccccccc}
    \toprule
    \multirow{2}{*}{Model}                        & \multicolumn{2}{c}{HTER (WMT20)}    & \multicolumn{6}{c}{HTER (WMT21)}                                                                                                                                                                                                                                        \\
    \cmidrule(lr){2-3} \cmidrule(lr){4-9}
                                                  & \multicolumn{1}{c}{\unidir{En}{De}} & \multicolumn{1}{c}{\unidir{En}{Zh}} & \multicolumn{1}{c}{\unidir{En}{De}} & \multicolumn{1}{c}{\unidir{En}{Zh}} & \multicolumn{1}{c}{\unidir{Ro}{En}} & \multicolumn{1}{c}{\unidir{Et}{En}} & \multicolumn{1}{c}{\unidir{Ne}{En}} & \multicolumn{1}{c}{\unidir{Si}{En}} \\
    \midrule
    LaBSE                                         & 0.104                               & 0.238                               & 0.119                               & 0.122                               & 0.748                               & 0.547                               & 0.451                               & 0.420                               \\
    MEAT                                          & 0.216                               & 0.384                               & 0.208                               & 0.215                               & 0.763                               & 0.614                               & 0.516                               & 0.501                               \\
    M2M-100 (w/o FT)                              & 0.323                               & 0.333                               & 0.269                               & 0.189                               & \textbf{0.808}                      & \textbf{0.679}                      & 0.248                               & 0.386                               \\
    M2M-100 (w/ FT)                               & 0.361                               & 0.340                               & 0.287                               & 0.200                               & 0.718                               & 0.674                               & 0.394                               & 0.389                               \\
    \midrule
    XLM-R$_\mathrm{Base}$+MLP (split)             & 0.393                               & 0.451$^\ast$                        & 0.280                               & 0.149                               & 0.347                               & 0.390                               & 0.391                               & 0.466                               \\
    XLM-R$_\mathrm{Large}$+MLP (split)            & 0.389                               & 0.457$^\ast$                        & 0.271                               & 0.182                               & 0.390                               & 0.446                               & 0.448                               & 0.503                               \\
    \textsc{InfoXLM}$_\mathrm{Base}$+MLP (split)  & 0.420$^\ast$                        & 0.452$^\ast$                        & 0.321                               & 0.178                               & 0.364                               & 0.411                               & 0.446                               & 0.478                               \\
    LaBSE+MLP (split)                             & 0.423$^\ast$                        & 0.466$^\ast$                        & 0.324                               & 0.194                               & 0.407                               & 0.432                               & 0.466                               & 0.527                               \\
    \midrule
    XLM-R$_\mathrm{Base}$+MLP (concat)            & 0.456$^\ast$                        & 0.492$^\ast$                        & 0.282                               & 0.180                               & 0.593                               & 0.480                               & 0.434                               & 0.536                               \\
    XLM-R$_\mathrm{Large}$+MLP (concat)           & 0.455$^\ast$                        & 0.542$^\ast$                        & 0.297                               & 0.214                               & 0.643                               & 0.583                               & \textbf{0.601}$^\ast$               & 0.608$^\ast$                        \\
    \textsc{InfoXLM}$_\mathrm{Base}$+MLP (concat) & \textbf{0.506}$^\ast$               & 0.524$^\ast$                        & \textbf{0.384}$^\ast$               & 0.212                               & 0.572                               & 0.530                               & 0.534                               & 0.591$^\ast$                        \\
    LaBSE+MLP (concat)                            & 0.483$^\ast$                        & \textbf{0.548}$^\ast$               & 0.352$^\ast$                        & \textbf{0.218}                      & 0.649                               & 0.569                               & 0.598$^\ast$                        & \textbf{0.611}$^\ast$               \\
    \bottomrule
  \end{tabular}
  \caption{Pearson's $r$ with HTER for non-zero-shot translation directions: \textbf{bold} and ``$\ast$'' respectively indicate the highest value in each column and a significant improvement over all the baselines.}
  \label{table:wmt-result}
  \bigskip
  \begin{tabular}{lccccc}
    \toprule
    \multirow{2}{*}{Model}                        & \multicolumn{5}{c}{HTER (WMT21)}                                                                                                                                                            \\
    \cmidrule(lr){2-6}
                                                  & \multicolumn{1}{c}{\unidir{Ru}{En}} & \multicolumn{1}{c}{\unidir{En}{Cs}} & \multicolumn{1}{c}{\unidir{En}{Ja}} & \multicolumn{1}{c}{\unidir{Km}{En}} & \multicolumn{1}{c}{\unidir{Ps}{En}} \\
    \midrule
    LaBSE                                         & 0.317                               & 0.065                               & 0.144                               & 0.395                               & 0.339                               \\
    MEAT                                          & \textbf{0.402}                      & 0.196                               & 0.198                               & 0.535                               & 0.424                               \\
    M2M-100 (w/o FT)                              & 0.339                               & 0.265                               & 0.154                               & 0.406                               & 0.336                               \\
    \midrule
    XLM-R$_\mathrm{Base}$+MLP (split)             & 0.228                               & 0.290                               & 0.151                               & 0.386                               & 0.233                               \\
    XLM-R$_\mathrm{Large}$+MLP (split)            & 0.315                               & 0.327$^\ast$                        & 0.173                               & 0.407                               & 0.237                               \\
    \textsc{InfoXLM}$_\mathrm{Base}$+MLP (split)  & 0.291                               & 0.330$^\ast$                        & 0.126                               & 0.401                               & 0.176                               \\
    LaBSE+MLP (split)                             & 0.245                               & 0.315                               & 0.138                               & 0.425                               & 0.219                               \\
    \midrule
    XLM-R$_\mathrm{Base}$+MLP (concat)            & 0.379                               & 0.298                               & 0.150                               & 0.507                               & 0.322                               \\
    XLM-R$_\mathrm{Large}$+MLP (concat)           & 0.398                               & 0.301                               & 0.145                               & 0.580$^\ast$                        & \textbf{0.457}                      \\
    \textsc{InfoXLM}$_\mathrm{Base}$+MLP (concat) & 0.362                               & \textbf{0.379}$^\ast$               & \textbf{0.207}                      & 0.566                               & 0.428                               \\
    LaBSE+MLP (concat)                            & 0.357                               & 0.370$^\ast$                        & 0.170                               & \textbf{0.584}$^\ast$               & 0.407                               \\
    \bottomrule
  \end{tabular}
  \caption{Pearson's $r$ with HTER for zero-shot translation directions: \textbf{bold} and ``$\ast$'' follow those in \Tab{table:wmt-result}.}
  \label{table:wmt-result-zero-shot}
\end{table*}
%

\paragraph{TQE models:}

As the pre-trained multilingual encoder, we compared \mbox{XLM-R$_{\mathrm{Base}}$}\footnote{\url{https://huggingface.co/xlm-roberta-base}} \citep{conneau-etal-2020-unsupervised}, \mbox{XLM-R$_{\mathrm{Large}}$}\footnote{\url{https://huggingface.co/xlm-roberta-large}}, \mbox{\textsc{InfoXLM}$_{\mathrm{Base}}$}\footnote{\url{https://huggingface.co/microsoft/infoxlm-base}} \citep{chi-etal-2021-infoxlm}, and LaBSE\footnote{\url{https://huggingface.co/sentence-transformers/LaBSE}} \citep{feng-etal-2022-language}.  The last two have optimized the embedding for [CLS] tokens during pre-training.
Table \ref{tab:model_param} shows the numbers of parameters of the examined TQE model.
We trained each TQE model with the AdamW optimizer \citep{loshchilov2018decoupled} ($\beta_{1}=0.9$, $\beta_{2}=0.999$, $\epsilon=1\times 10^{-8}$) on batches of 128 tuples of synthetic TQE data for all the six translation directions, also updating the parameters of the pre-trained multilingual encoder.
We performed early stopping for training with a patience of $3$, according to a validation loss calculated for every $10$k steps on a randomly sub-sampled $10\%$ tuples of training data.
We separately trained four models with different learning rates of $5\times 10^{-5}$, $1\times 10^{-5}$, $5\times 10^{-6}$, and $1\times 10^{-6}$, and selected the best model for evaluation according to the validation loss.

\begin{table*}[p]
  \centering
  \small
  \begin{tabular}{lccccccccc}
    \toprule
    \multirow{2}{*}{Model}               & \multicolumn{6}{c}{Direct Assessment (WMT20)}                                                                                                                                                                                               \\
    \cmidrule(lr){2-7}
                                         & \multicolumn{1}{c}{\unidir{En}{De}}           & \multicolumn{1}{c}{\unidir{En}{Zh}} & \multicolumn{1}{c}{\unidir{Ro}{En}} & \multicolumn{1}{c}{\unidir{Et}{En}} & \multicolumn{1}{c}{\unidir{Ne}{En}} & \multicolumn{1}{c}{\unidir{Si}{En}} \\
    \midrule
    LaBSE                                & 0.084                                         & 0.036                               & 0.705                               & 0.550                               & 0.547                               & 0.455                               \\
    MEAT                                 & 0.213                                         & 0.223                               & 0.718                               & 0.585                               & \textbf{0.633}                      & \textbf{0.570}                      \\
    M2M-100 (w/o FT)                     & 0.246                                         & 0.209                               & \textbf{0.807}                      & 0.669                               & 0.298                               & 0.398                               \\
    M2M-100 (w/ FT)                      & 0.254                                         & 0.279                               & 0.780                               & \textbf{0.680}                      & 0.497                               & 0.418                               \\
    \midrule
    XLM-R$_\mathrm{Base}$+MLP (split)    & 0.230                                         & 0.218                               & 0.329                               & 0.367                               & 0.386                               & 0.408                               \\
    XLM-R$_\mathrm{Large}$+MLP (split)   & 0.208                                         & 0.231                               & 0.407                               & 0.436                               & 0.392                               & 0.427                               \\
    InfoXLM$_\mathrm{Base}$+MLP (split)  & 0.242                                         & 0.238                               & 0.362                               & 0.437                               & 0.410                               & 0.419                               \\
    LaBSE+MLP (split)                    & 0.219                                         & 0.217                               & 0.386                               & 0.455                               & 0.420                               & 0.421                               \\
    \midrule
    XLM-R$_\mathrm{Base}$+MLP (concat)   & 0.231                                         & 0.236                               & 0.535                               & 0.433                               & 0.430                               & 0.462                               \\
    XLM-R$_\mathrm{Large}$+MLP (concat)  & 0.183                                         & 0.256                               & 0.590                               & 0.526                               & 0.541                               & 0.479                               \\
    InfoXLM$_\mathrm{Base}$+MLP (concat) & \textbf{0.286}                                & \textbf{0.285}                      & 0.502                               & 0.486                               & 0.524                               & 0.491                               \\
    LaBSE+MLP (concat)                   & 0.252                                         & 0.264                               & 0.569                               & 0.527                               & 0.605                               & 0.510                               \\
    \bottomrule
  \end{tabular}
  \caption{Pearson's $r$ with DA score for non-zero-shot translation directions: \textbf{bold} and ``$\ast$'' follow those in \Tab{table:wmt-result}.}
  \label{table:wmt20-da-results}
  \bigskip
  \centering
  \small
  \begin{tabular}{lccccccccc}
    \toprule
    \multirow{2}{*}{Model}                        & \multicolumn{6}{c}{Direct Assessment (WMT21)}                                                                                                                                                                                               \\
    \cmidrule(lr){2-7}
                                                  & \multicolumn{1}{c}{\unidir{En}{De}}           & \multicolumn{1}{c}{\unidir{En}{Zh}} & \multicolumn{1}{c}{\unidir{Ro}{En}} & \multicolumn{1}{c}{\unidir{Et}{En}} & \multicolumn{1}{c}{\unidir{Ne}{En}} & \multicolumn{1}{c}{\unidir{Si}{En}} \\
    \midrule
    LaBSE                                         & 0.089                                         & 0.099                               & 0.718                               & 0.591                               & 0.585                               & 0.360                               \\
    MEAT                                          & 0.176                                         & 0.270                               & 0.722                               & 0.607                               & \textbf{0.670}                      & 0.459                               \\
    M2M-100 (w/o FT)                              & 0.212                                         & 0.250                               & \textbf{0.805}                      & \textbf{0.652}                      & 0.357                               & 0.384                               \\
    M2M-100 (w/ FT)                               & 0.211                                         & 0.321                               & 0.767                               & 0.651                               & 0.534                               & 0.367                               \\
    \midrule
    XLM-R$_\mathrm{Base}$+MLP (split)             & 0.187                                         & 0.320                               & 0.373                               & 0.348                               & 0.489                               & 0.375                               \\
    XLM-R$_\mathrm{Large}$+MLP (split)            & 0.172                                         & 0.360                               & 0.429                               & 0.416                               & 0.508                               & 0.392                               \\
    InfoXLM$_\mathrm{Base}$+MLP (split)           & 0.226                                         & 0.333                               & 0.388                               & 0.364                               & 0.521                               & 0.392                               \\
    LaBSE+MLP (split)                             & 0.211                                         & 0.340                               & 0.444                               & 0.404                               & 0.547                               & 0.386                               \\
    \midrule
    XLM-R$_\mathrm{Base}$+MLP (concat)            & 0.195                                         & 0.364                               & 0.578                               & 0.429                               & 0.478                               & 0.418                               \\
    XLM-R$_\mathrm{Large}$+MLP (concat)           & 0.153                                         & 0.370$^\ast$                        & 0.628                               & 0.510                               & 0.626                               & 0.431                               \\
    \textsc{InfoXLM}$_\mathrm{Base}$+MLP (concat) & \textbf{0.265}$^\ast$                         & \textbf{0.385}$^\ast$               & 0.539                               & 0.447                               & 0.586                               & 0.450                               \\
    LaBSE+MLP (concat)                            & 0.259                                         & 0.382$^\ast$                        & 0.625                               & 0.514                               & 0.669                               & \textbf{0.465}                      \\
    \bottomrule
  \end{tabular}
  \caption{Pearson's $r$ with DA score for non-zero-shot translation directions: \textbf{bold} and ``$\ast$'' follow those in \Tab{table:wmt-result}.}
  \label{table:wmt21-da-results}
  \bigskip
  \centering
  \small
  \begin{tabular}{lccccc}
    \toprule
    \multirow{2}{*}{Model}               & \multicolumn{5}{c}{Direct Assessment (WMT21)}                                                                                                                                                         \\
    \cmidrule(lr){2-6}
                                         & \multicolumn{1}{c}{\unidir{Ru}{En}}           & \multicolumn{1}{c}{\unidir{En}{Cs}} & \multicolumn{1}{c}{\unidir{En}{Ja}} & \multicolumn{1}{c}{\unidir{Km}{En}} & \multicolumn{1}{c}{\unidir{Ps}{En}} \\
    \midrule
    LaBSE                                & 0.387                                         & 0.131                               & 0.126                               & 0.357                               & 0.363                               \\
    MEAT                                 & \textbf{0.524}                                & 0.269                               & 0.181                               & \textbf{0.531}                      & \textbf{0.458}                      \\
    M2M-100 (w/o FT)                     & 0.410                                         & \textbf{0.350}                      & 0.121                               & 0.349                               & 0.388                               \\
    \midrule
    XLM-R$_\mathrm{Base}$+MLP (split)    & 0.335                                         & 0.230                               & 0.135                               & 0.371                               & 0.232                               \\
    XLM-R$_\mathrm{Large}$+MLP (split)   & 0.400                                         & 0.253                               & \textbf{0.189}                      & 0.402                               & 0.248                               \\
    InfoXLM$_\mathrm{Base}$+MLP (split)  & 0.361                                         & 0.237                               & 0.173                               & 0.387                               & 0.194                               \\
    LaBSE+MLP (split)                    & 0.335                                         & 0.248                               & 0.139                               & 0.414                               & 0.207                               \\
    \midrule
    XLM-R$_\mathrm{Base}$+MLP (concat)   & 0.433                                         & 0.238                               & 0.130                               & 0.487                               & 0.338                               \\
    XLM-R$_\mathrm{Large}$+MLP (concat)  & 0.466                                         & 0.240                               & 0.155                               & 0.508                               & 0.455                               \\
    InfoXLM$_\mathrm{Base}$+MLP (concat) & 0.392                                         & 0.312                               & 0.166                               & 0.504                               & 0.434                               \\
    LaBSE+MLP (concat)                   & 0.425                                         & 0.324                               & 0.179                               & 0.520                               & 0.412                               \\
    \bottomrule
  \end{tabular}
  \caption{Pearson's $r$ with DA score for zero-shot translation directions: \textbf{bold} and ``$\ast$'' follow those in \Tab{table:wmt-result}.}
  \label{table:wmt21-da-result-zero-shot}
\end{table*}
\begin{table*}[t]
  \centering
  \small
  \begin{tabular}{l@{}c@{}cccccccc}
    \toprule
    \multirow{2}{*}{Model}                        & \multirow{2}{*}{\#TD} & \multicolumn{2}{c}{HTER (WMT20)}    & \multicolumn{6}{c}{HTER (WMT21)}                                                                                                                                                                                                                                        \\
    \cmidrule(lr){3-4} \cmidrule(lr){5-10}
                                                  &                       & \multicolumn{1}{c}{\unidir{En}{De}} & \multicolumn{1}{c}{\unidir{En}{Zh}} & \multicolumn{1}{c}{\unidir{En}{De}} & \multicolumn{1}{c}{\unidir{En}{Zh}} & \multicolumn{1}{c}{\unidir{Ro}{En}} & \multicolumn{1}{c}{\unidir{Et}{En}} & \multicolumn{1}{c}{\unidir{Ne}{En}} & \multicolumn{1}{c}{\unidir{Si}{En}} \\
    \midrule
    \textsc{InfoXLM}$_\mathrm{Base}$+MLP (concat) & 6                     & \textbf{0.506}$^\ast$               & \textbf{0.524}$^\ast$               & \textbf{0.384}$^\ast$               & \textbf{0.212}$^\ast$               & \textbf{0.572}$^\ast$               & \textbf{0.530}$^\ast$               & \textbf{0.534}$^\ast$               & 0.591                               \\
    \textsc{InfoXLM}$_\mathrm{Base}$+MLP (concat) & 1                     & 0.407                               & 0.435                               & 0.284                               & 0.155                               & 0.491                               & 0.296                               & 0.341                               & \textbf{0.593}                      \\
    \midrule
    LaBSE+MLP (concat)                            & 6                     & \textbf{0.483}                      & \textbf{0.548}$^\ast$               & \textbf{0.352}                      & \textbf{0.218}$^\ast$               & \textbf{0.649}$^\ast$               & 0.569                               & \textbf{0.598}$^\ast$               & \textbf{0.611}                      \\
    LaBSE+MLP (concat)                            & 1                     & 0.466                               & 0.475                               & 0.344                               & 0.184                               & 0.511                               & \textbf{0.602}$^\ast$               & 0.268                               & 0.586                               \\
    \bottomrule
  \end{tabular}
  \caption{Pearson's $r$ with HTER of LaBSE+MLP and \textsc{InfoXLM}$_\mathrm{Base}$+MLP trained on different numbers of translation directions (\#TD): \textbf{bold} indicates the winner and ``$\ast$'' shows the value is significantly higher than the other.}
  \label{table:wmt-result-analysis}
\end{table*}
%

\paragraph{Baselines:}

Our baselines include three existing unsupervised TQE methods (four models) that do not use synthetic TQE data but bilingual parallel data.
LaBSE is the one explained in \Sec{sec:previous}.
While MEAT\footnote{\url{https://github.com/kuro961/MEAT}} was trained on a smallish parallel corpus in the original work \citep{kuroda-etal-2022-adversarial}, we trained it on the entire bilingual corpora in \Tab{table:num-data}.
As the forced-decoding methods, we compared the original and fine-tuned M2M-100 models.
The latter, denoted as ``w/ FT,'' were six direction-specific models that are also used for generating synthetic TQE data.

\paragraph{Evaluation metric:}

Accuracy of each TQE model was measured with Pearson's product-moment correlation coefficient (Pearson's $r$) between its outputs and gold scores.
The baseline methods are designed to give a higher score for a better MT output, unlike HTER.
Therefore, we multiplied $-1$ to all their predicted scores before computing Pearson's $r$ with the gold HTER.
We did the same for the predicted scores of the models trained on synthetic TQE data before computing Pearson's $r$ with the DA score.
Statistical significance between coefficients ($p<0.05$) was tested with Williams significance test
using \texttt{mt-qe-eval}.\footnote{\url{https://github.com/ygraham/mt-qe-eval}}

\subsection{Results on Predicting HTER}

\Tab{table:wmt-result} shows our experimental results on the non-zero-shot translation directions.
Some neural TQE models with the ``concat'' setting outperformed the baselines on the high- and low-resource translation directions.
When compared the two variants based on the same pre-trained models, the ``concat'' model was consistently superior to the ``split'' model, suggesting the usefulness of capturing cross-lingual token-level correspondences during encoding, as demonstrated in \citet{ranasinghe-etal-2020-transquest}.
\textsc{InfoXLM}$_\mathrm{Base}$ consistently outperformed XLM-R$_\mathrm{Base}$.
Considering that both have the same number of parameters and identical vocabulary (see \Tab{tab:model_param}), this indicates that the [CLS] embedding optimized during pre-training \textsc{InfoXLM}$_\mathrm{Base}$ helps TQE.
We consider that LaBSE also benefits from the optimized embedding for [CLS].
For most test sets, the larger XLM-R (XLM-R$_\mathrm{Large}$+MLP) outperformed its smaller counterpart (XLM-R$_\mathrm{Base}$+MLP), thanks to the larger number of parameters.

TQE models trained on synthetic data show deteriorated results in the medium-resource translation directions, i.e., \unidir{Ro}{En} and \unidir{Et}{En}.\footnote{Previous work \citep{tiyajamorn-etal-2021-language,kuroda-etal-2022-adversarial} on predicting DA scores also shows a smaller improvement on these two translation directions compared to others.}
Similarly, the performance of fine-tuned M2M-100 was worse than the original M2M-100 on these two translation directions.
To delve the reasons for it, we need to untangle the following possible factors: (a) size of synthetic training data, (b) language-specific features, (c) domain mismatch between parallel data and test data, (d) differences in the quality of the hypothesis, and (e) differences in post-edit quality.

\Tab{table:wmt-result-zero-shot} shows the results on the zero-shot translation directions.
The ``concat'' version of TQE models outperformed baselines on \unidir{En}{CS} and \unidir{Km}{En},
indicating that synthetic data also help zero-shot TQE depending on translation direction.
Similarly to the results in \Tab{table:wmt-result}, we observed that the following factors affect performance: (a) encoding methods (``split'' models vs. ``concat'' models), (b) optimization of [CLS] during pre-training (XLM-R$_\mathrm{Base}$+MLP (concat) vs. \textsc{InfoXLM}$_\mathrm{Base}$+MLP (concat)), and (c) the number of parameters (XLM-R$_\mathrm{Base}$+MLP (concat) vs. XLM-R$_\mathrm{Large}$+MLP (concat)).

\subsection{Results on Predicting DA Scores}

Preceding sections have focused on HTER as the target metric to predict, assuming that labels of synthetic data, i.e., TER, should be correlated with HTER.
To investigate performance in another evaluation metric that is less correlated with HTER, we also evaluated TQE models on predicting DA scores.

Tables~\ref{table:wmt20-da-results} and \ref{table:wmt21-da-results} show experimental results on six translation directions in the WMT20 and WMT21 TQE Task 1 test sets, respectively.
\textsc{InfoXLM}$_\mathrm{Base}$+MLP (concat) predicted DA scores on high-resource translation directions better than all the baseline models.
In contrast, for middle-resource and low-resource translation directions, none of the neural TQE models trained on synthetic data outperformed MEAT, except for LaBSE+MLP (concat) for \unidir{Si}{En} in WMT21 test set.

\Tab{table:wmt21-da-result-zero-shot} shows the results on the zero-shot translation directions.
XLM-R$_\mathrm{Large}$+MLP (concat) and LaBSE+MLP (concat) outperformed two of three baselines in all the translation directions except for \unidir{En}{Cs}, but did not outperform MEAT.

These results demonstrate that exploiting parallel corpora, even for other language pairs, in the way of what MEAT does, contributed to the prediction of DA score more than synthetic data labeled with TER scores.  We speculate that this stems from the low reliability of TER score computed from separately produced MT output and reference translation and the low relevance between TER and DA score.

\section{Impact of Multi-directional Training}

Previous work has revealed that neural language models and NMT models can benefit from training data for multiple languages and multiple translation directions \citep{conneau-etal-2020-unsupervised, tang-etal-2021-multilingual}.

Our investigation demonstrates that it also applies to TQE models, as shown in \Tab{table:wmt-result-analysis}.
We took LaBSE+MLP (concat) and \textsc{InfoXLM}$_\mathrm{Base}$+MLP (concat) according to their superior results in \Tab{table:wmt-result}, compared two variants of these models trained exclusively on synthetic TQE data for multiple (\#TD=6, \Sec{sec:experiment}) or single (\#TD=1, \citep{fomicheva-etal-2020-unsupervised,tuan-etal-2021-quality,eo-etal-2021-dealing}) translation directions on the same TQE task as in \Sec{sec:experiment}, and confirmed that the former outperformed the latter on most translation directions.

\section{Conclusion}

In this paper, we investigated the usefulness of synthetic training data for TQE, in particular for predicting HTER using pre-trained multilingual encoders.
Our experiment on WMT20 and WMT21 TQE datasets confirmed that the TQE models trained on synthetic TQE data better predict HTER on high- and low-resource translation directions, compared to existing unsupervised TQE methods that do not use synthetic TQE data.

Our work has the following contributions that have never been examined in previous work: (a) we found pre-trained multilingual encoders with optimized embedding for [CLS], i.e., LaBSE and \textsc{InfoXLM}, can be trained well exclusively on synthetic TQE data, (b) we reconfirmed that joint encoding of source sentence and MT output works better than separate encoding as in the supervised setting \citep{ranasinghe-etal-2020-transquest}, (c) we identified the positive impact of using multi-directional synthetic TQE data in unsupervised TQE training, and (d) synthetic data with TER are also useful for predicting DA score on high- and low-resource translation directions.

\section*{Limitations}

Our experiment covers only eleven translation directions,
and our results do not guarantee the same conclusions on other translation directions.
As demonstrated in our experiment, even for the same translation direction (two \unidir{En}{De} and two \unidir{En}{Zh} tasks in WMT20 and WMT21),
the TQE accuracy can be visibly different.  This implies that the difficulty of the task depends on the characteristics of the test data, the MT systems used for generating MT outputs,
human annotators recruited for obtaining gold HTER, and so forth.

The entire experiment was carried out on a single V100 GPU.
If we had a more powerful environment, higher TQE accuracy could be achieved by employing larger pre-trained multilingual encoders, larger batch sizes, longer training, and so forth.

\section*{Ethics Statement}

As shown in our experiment, predicted scores does not perfectly
correlate with the gold score obtained through human post-editing.
Therefore, such predicted scores could mislead potential users.
Note that this is not specific to our work, but common in the TQE task.


\section*{Acknowledgments}
This work was supported by JSPS KAKENHI Grant Number JP20K19861.
These research results were obtained from the commissioned research (No.22501) by National Institute of Information and Communications Technology (NICT), Japan.

\bibliography{custom}

\begin{thebibliography}{30}
\expandafter\ifx\csname natexlab\endcsname\relax\def\natexlab#1{#1}\fi

\bibitem[{Blatz et~al.(2004)Blatz, Fitzgerald, Foster, Gandrabur, Goutte, Kulesza, Sanchis, and Ueffing}]{blatz-etal-2004-confidence}
John Blatz, Erin Fitzgerald, George Foster, Simona Gandrabur, Cyril Goutte, Alex Kulesza, Alberto Sanchis, and Nicola Ueffing. 2004.
\newblock \href {https://aclanthology.org/C04-1046} {{Confidence Estimation for Machine Translation}}.
\newblock In \emph{Proceedings of the 20th International Conference on Computational Linguistics}, pages 315--321.

\bibitem[{Chi et~al.(2021)Chi, Dong, Wei, Yang, Singhal, Wang, Song, Mao, Huang, and Zhou}]{chi-etal-2021-infoxlm}
Zewen Chi, Li~Dong, Furu Wei, Nan Yang, Saksham Singhal, Wenhui Wang, Xia Song, Xian-Ling Mao, Heyan Huang, and Ming Zhou. 2021.
\newblock \href {https://doi.org/10.18653/v1/2021.naacl-main.280} {{InfoXLM: An Information-Theoretic Framework for Cross-Lingual Language Model Pre-Training}}.
\newblock In \emph{Proceedings of the 2021 Conference of the North American Chapter of the Association for Computational Linguistics: Human Language Technologies}, pages 3576--3588.

\bibitem[{Conneau et~al.(2020)Conneau, Khandelwal, Goyal, Chaudhary, Wenzek, Guzm{\'a}n, Grave, Ott, Zettlemoyer, and Stoyanov}]{conneau-etal-2020-unsupervised}
Alexis Conneau, Kartikay Khandelwal, Naman Goyal, Vishrav Chaudhary, Guillaume Wenzek, Francisco Guzm{\'a}n, Edouard Grave, Myle Ott, Luke Zettlemoyer, and Veselin Stoyanov. 2020.
\newblock \href {https://aclanthology.org/2020.acl-main.747} {{Unsupervised Cross-lingual Representation Learning at Scale}}.
\newblock In \emph{Proceedings of the 58th Annual Meeting of the Association for Computational Linguistics}, pages 8440--8451.

\bibitem[{Devlin et~al.(2019)Devlin, Chang, Lee, and Toutanova}]{devlin-etal-2019-bert}
Jacob Devlin, Ming-Wei Chang, Kenton Lee, and Kristina Toutanova. 2019.
\newblock \href {https://aclanthology.org/N19-1423} {{BERT: Pre-training of Deep Bidirectional Transformers for Language Understanding}}.
\newblock In \emph{Proceedings of the 2019 Conference of the North {A}merican Chapter of the Association for Computational Linguistics: Human Language Technologies, Volume 1 (Long and Short Papers)}, pages 4171--4186.

\bibitem[{Eo et~al.(2021)Eo, Park, Moon, Seo, and Lim}]{eo-etal-2021-dealing}
Sugyeong Eo, Chanjun Park, Hyeonseok Moon, Jaehyung Seo, and Heuiseok Lim. 2021.
\newblock \href {https://aclanthology.org/2021.mtsummit-loresmt.1} {{Dealing with the Paradox of Quality Estimation}}.
\newblock In \emph{Proceedings of the 4th Workshop on Technologies for MT of Low Resource Languages (LoResMT2021)}, pages 1--10.

\bibitem[{Fan et~al.(2022)Fan, Bhosale, Schwenk, Ma, El-Kishky, Goyal, Baines, Celebi, Wenzek, Chaudhary, Goyal, Birch, Liptchinsky, Edunov, Grave, Auli, and Joulin}]{fan-etal-2022-beyond}
Angela Fan, Shruti Bhosale, Holger Schwenk, Zhiyi Ma, Ahmed El-Kishky, Siddharth Goyal, Mandeep Baines, Onur Celebi, Guillaume Wenzek, Vishrav Chaudhary, Naman Goyal, Tom Birch, Vitaliy Liptchinsky, Sergey Edunov, Edouard Grave, Michael Auli, and Armand Joulin. 2022.
\newblock \href {https://www.jmlr.org/papers/volume22/20-1307/20-1307.pdf} {{Beyond English-Centric Multilingual Machine Translation}}.
\newblock \emph{Journal of Machine Learning Research}, 22(1):4839--4886.

\bibitem[{Feng et~al.(2022)Feng, Yang, Cer, Arivazhagan, and Wang}]{feng-etal-2022-language}
Fangxiaoyu Feng, Yinfei Yang, Daniel Cer, Naveen Arivazhagan, and Wei Wang. 2022.
\newblock \href {https://aclanthology.org/2022.acl-long.62} {{Language-agnostic {BERT} Sentence Embedding}}.
\newblock In \emph{Proceedings of the 60th Annual Meeting of the Association for Computational Linguistics}, pages 878--891.

\bibitem[{Fomicheva et~al.(2022)Fomicheva, Sun, Fonseca, Zerva, Blain, Chaudhary, Guzm{\'a}n, Lopatina, Specia, and Martins}]{fomicheva-etal-2022-mlqe}
Marina Fomicheva, Shuo Sun, Erick Fonseca, Chrysoula Zerva, Fr{\'e}d{\'e}ric Blain, Vishrav Chaudhary, Francisco Guzm{\'a}n, Nina Lopatina, Lucia Specia, and Andr{\'e} F.~T. Martins. 2022.
\newblock \href {https://aclanthology.org/2022.lrec-1.530} {{MLQE-PE: A Multilingual Quality Estimation and Post-Editing Dataset}}.
\newblock In \emph{Proceedings of the Thirteenth Language Resources and Evaluation Conference}, pages 4963--4974.

\bibitem[{Fomicheva et~al.(2020)Fomicheva, Sun, Yankovskaya, Blain, Guzm{\'a}n, Fishel, Aletras, Chaudhary, and Specia}]{fomicheva-etal-2020-unsupervised}
Marina Fomicheva, Shuo Sun, Lisa Yankovskaya, Fr{\'e}d{\'e}ric Blain, Francisco Guzm{\'a}n, Mark Fishel, Nikolaos Aletras, Vishrav Chaudhary, and Lucia Specia. 2020.
\newblock \href {https://doi.org/10.1162/tacl_a_00330} {{Unsupervised Quality Estimation for Neural Machine Translation}}.
\newblock \emph{Transactions of the Association for Computational Linguistics}, 8:539--555.

\bibitem[{Fonseca et~al.(2019)Fonseca, Yankovskaya, Martins, Fishel, and Federmann}]{fonseca-etal-2019-findings}
Erick Fonseca, Lisa Yankovskaya, Andr{\'e} F.~T. Martins, Mark Fishel, and Christian Federmann. 2019.
\newblock \href {https://doi.org/10.18653/v1/W19-5401} {{Findings of the WMT 2019 Shared Tasks on Quality Estimation}}.
\newblock In \emph{Proceedings of the Fourth Conference on Machine Translation (Volume 3: Shared Task Papers, Day 2)}, pages 1--10.

\bibitem[{ISO/TC37(2017)}]{ISO18587:17}
ISO/TC37. 2017.
\newblock \href {https://iso.org/standard/62970.html} {{ISO} 18587:2017 translation services: Post-editing of machine translation output: Requirements}.

\bibitem[{Kim et~al.(2017)Kim, Lee, and Na}]{kim-etal-2017-predictor}
Hyun Kim, Jong-Hyeok Lee, and Seung-Hoon Na. 2017.
\newblock \href {https://doi.org/10.18653/v1/W17-4763} {{Predictor-Estimator using Multilevel Task Learning with Stack Propagation for Neural Quality Estimation}}.
\newblock In \emph{Proceedings of the Second Conference on Machine Translation}, pages 562--568.

\bibitem[{Kuroda et~al.(2022)Kuroda, Kajiwara, Arase, and Ninomiya}]{kuroda-etal-2022-adversarial}
Yuto Kuroda, Tomoyuki Kajiwara, Yuki Arase, and Takashi Ninomiya. 2022.
\newblock \href {https://aclanthology.org/2022.coling-1.465} {{Adversarial Training on Disentangling Meaning and Language Representations for Unsupervised Quality Estimation}}.
\newblock In \emph{Proceedings of the 29th International Conference on Computational Linguistics}, pages 5240--5245.

\bibitem[{Loshchilov and Hutter(2019)}]{loshchilov2018decoupled}
Ilya Loshchilov and Frank Hutter. 2019.
\newblock \href {https://openreview.net/forum?id=Bkg6RiCqY7} {{Decoupled Weight Decay Regularization}}.
\newblock In \emph{Proceedings of the 7th International Conference on Learning Representations}.

\bibitem[{Ott et~al.(2019)Ott, Edunov, Baevski, Fan, Gross, Ng, Grangier, and Auli}]{ott-2019}
Myle Ott, Sergey Edunov, Alexei Baevski, Angela Fan, Sam Gross, Nathan Ng, David Grangier, and Michael Auli. 2019.
\newblock \href {https://aclanthology.org/N19-4009} {{fairseq: A Fast, Extensible Toolkit for Sequence Modeling}}.
\newblock In \emph{Proceedings of the 2019 Conference of the North {A}merican Chapter of the Association for Computational Linguistics (Demonstrations)}, pages 48--53.

\bibitem[{Post(2018)}]{post-2018-call}
Matt Post. 2018.
\newblock \href {https://www.aclweb.org/anthology/W18-6319} {{A Call for Clarity in Reporting BLEU Scores}}.
\newblock In \emph{Proceedings of the Third Conference on Machine Translation: Research Papers}, pages 186--191.

\bibitem[{Ranasinghe et~al.(2020)Ranasinghe, Orasan, and Mitkov}]{ranasinghe-etal-2020-transquest}
Tharindu Ranasinghe, Constantin Orasan, and Ruslan Mitkov. 2020.
\newblock \href {https://aclanthology.org/2020.coling-main.445} {{TransQuest: Translation Quality Estimation with Cross-lingual Transformers}}.
\newblock In \emph{Proceedings of the 28th International Conference on Computational Linguistics}, pages 5070--5081.

\bibitem[{Rei et~al.(2022)Rei, Treviso, Guerreiro, Zerva, Farinha, Maroti, C.~de Souza, Glushkova, Alves, Coheur, Lavie, and Martins}]{rei-etal-2022-cometkiwi}
Ricardo Rei, Marcos Treviso, Nuno~M. Guerreiro, Chrysoula Zerva, Ana~C Farinha, Christine Maroti, Jos{\'e}~G. C.~de Souza, Taisiya Glushkova, Duarte Alves, Luisa Coheur, Alon Lavie, and Andr{\'e} F.~T. Martins. 2022.
\newblock \href {https://aclanthology.org/2022.wmt-1.60} {{{C}omet{K}iwi: {IST}-Unbabel 2022 Submission for the Quality Estimation Shared Task}}.
\newblock In \emph{Proceedings of the Seventh Conference on Machine Translation (WMT)}, pages 634--645.

\bibitem[{Snover et~al.(2006)Snover, Dorr, Schwartz, Micciulla, and Makhoul}]{snover-etal-2006-study}
Matthew Snover, Bonnie Dorr, Rich Schwartz, Linnea Micciulla, and John Makhoul. 2006.
\newblock \href {https://aclanthology.org/2006.amta-papers.25} {{A Study of Translation Edit Rate with Targeted Human Annotation}}.
\newblock In \emph{Proceedings of the 7th Conference of the Association for Machine Translation in the Americas: Technical Papers}, pages 223--231.

\bibitem[{Song et~al.(2021)Song, Zhao, and Specia}]{song-etal-2021-sentsim}
Yurun Song, Junchen Zhao, and Lucia Specia. 2021.
\newblock \href {https://doi.org/10.18653/v1/2021.naacl-main.252} {{SentSim: Crosslingual Semantic Evaluation of Machine Translation}}.
\newblock In \emph{Proceedings of the 2021 Conference of the North American Chapter of the Association for Computational Linguistics: Human Language Technologies}, pages 3143--3156.

\bibitem[{Specia et~al.(2020)Specia, Blain, Fomicheva, Fonseca, Chaudhary, Guzm{\'a}n, and Martins}]{specia-etal-2020-findings-wmt}
Lucia Specia, Fr{\'e}d{\'e}ric Blain, Marina Fomicheva, Erick Fonseca, Vishrav Chaudhary, Francisco Guzm{\'a}n, and Andr{\'e} F.~T. Martins. 2020.
\newblock \href {https://aclanthology.org/2020.wmt-1.79} {{Findings of the {WMT} 2020 Shared Task on Quality Estimation}}.
\newblock In \emph{Proceedings of the Fifth Conference on Machine Translation}, pages 743--764.

\bibitem[{Specia et~al.(2021)Specia, Blain, Fomicheva, Zerva, Li, Chaudhary, and Martins}]{specia-etal-2021-findings}
Lucia Specia, Fr{\'e}d{\'e}ric Blain, Marina Fomicheva, Chrysoula Zerva, Zhenhao Li, Vishrav Chaudhary, and Andr{\'e} F.~T. Martins. 2021.
\newblock \href {https://aclanthology.org/2021.wmt-1.71} {{Findings of the WMT 2021 Shared Task on Quality Estimation}}.
\newblock In \emph{Proceedings of the Sixth Conference on Machine Translation}, pages 684--725.

\bibitem[{Specia et~al.(2018)Specia, Scarton, and Paetzold}]{specia-2018}
Lucia Specia, Carolina Scarton, and Gustavo~Henrique Paetzold. 2018.
\newblock \href {https://doi.org/10.2200/S00854ED1V01Y201805HLT039} {{Quality Estimation for Machine Translation}}.
\newblock \emph{Synthesis Lectures on Human Language Technologies}, 11(1):1--162.

\bibitem[{Tang et~al.(2021)Tang, Tran, Li, Chen, Goyal, Chaudhary, Gu, and Fan}]{tang-etal-2021-multilingual}
Yuqing Tang, Chau Tran, Xian Li, Peng-Jen Chen, Naman Goyal, Vishrav Chaudhary, Jiatao Gu, and Angela Fan. 2021.
\newblock \href {https://doi.org/10.18653/v1/2021.findings-acl.304} {{Multilingual Translation from Denoising Pre-Training}}.
\newblock In \emph{Findings of the Association for Computational Linguistics: ACL-IJCNLP 2021}, pages 3450--3466.

\bibitem[{Thompson and Post(2020)}]{thompson-post-2020-automatic}
Brian Thompson and Matt Post. 2020.
\newblock \href {https://aclanthology.org/2020.emnlp-main.8} {{Automatic Machine Translation Evaluation in Many Languages via Zero-Shot Paraphrasing}}.
\newblock In \emph{Proceedings of the 2020 Conference on Empirical Methods in Natural Language Processing}, pages 90--121.

\bibitem[{Tiyajamorn et~al.(2021)Tiyajamorn, Kajiwara, Arase, and Onizuka}]{tiyajamorn-etal-2021-language}
Nattapong Tiyajamorn, Tomoyuki Kajiwara, Yuki Arase, and Makoto Onizuka. 2021.
\newblock \href {https://aclanthology.org/2021.emnlp-main.612} {{Language-agnostic Representation from Multilingual Sentence Encoders for Cross-lingual Similarity Estimation}}.
\newblock In \emph{Proceedings of the 2021 Conference on Empirical Methods in Natural Language Processing}, pages 7764--7774.

\bibitem[{Tuan et~al.(2021)Tuan, El-Kishky, Renduchintala, Chaudhary, Guzm{\'a}n, and Specia}]{tuan-etal-2021-quality}
Yi-Lin Tuan, Ahmed El-Kishky, Adithya Renduchintala, Vishrav Chaudhary, Francisco Guzm{\'a}n, and Lucia Specia. 2021.
\newblock \href {https://doi.org/10.18653/v1/2021.eacl-main.50} {{Quality Estimation without Human-labeled Data}}.
\newblock In \emph{Proceedings of the 16th Conference of the European Chapter of the Association for Computational Linguistics: Main Volume}, pages 619--625.

\bibitem[{Vaswani et~al.(2017)Vaswani, Shazeer, Parmar, Uszkoreit, Jones, Gomez, Kaiser, and Polosukhin}]{vaswani-2017}
Ashish Vaswani, Noam Shazeer, Niki Parmar, Jakob Uszkoreit, Llion Jones, Aidan~N. Gomez, undefinedukasz Kaiser, and Illia Polosukhin. 2017.
\newblock \href {https://papers.nips.cc/paper/2017/file/3f5ee243547dee91fbd053c1c4a845aa-Paper.pdf} {{Attention is All You Need}}.
\newblock In \emph{Proceedings of the 31st Conference on Neural Information Processing Systems}, pages 5998--6008.

\bibitem[{Wang et~al.(2021)Wang, Wang, Chen, Zhao, Luo, and Zhang}]{wang-etal-2021-qemind}
Jiayi Wang, Ke~Wang, Boxing Chen, Yu~Zhao, Weihua Luo, and Yuqi Zhang. 2021.
\newblock \href {https://aclanthology.org/2021.wmt-1.100} {{QEMind: Alibaba's Submission to the WMT21 Quality Estimation Shared Task}}.
\newblock In \emph{Proceedings of the Sixth Conference on Machine Translation}, pages 948--954.

\bibitem[{Zerva et~al.(2022)Zerva, Blain, Rei, Lertvittayakumjorn, C.~de Souza, Eger, Kanojia, Alves, Or{\u{a}}san, Fomicheva, Martins, and Specia}]{zerva-etal-2022-findings}
Chrysoula Zerva, Fr{\'e}d{\'e}ric Blain, Ricardo Rei, Piyawat Lertvittayakumjorn, Jos{\'e}~G. C.~de Souza, Steffen Eger, Diptesh Kanojia, Duarte Alves, Constantin Or{\u{a}}san, Marina Fomicheva, Andr{\'e} F.~T. Martins, and Lucia Specia. 2022.
\newblock \href {https://aclanthology.org/2022.wmt-1.3} {{Findings of the {WMT} 2022 Shared Task on Quality Estimation}}.
\newblock In \emph{Proceedings of the Seventh Conference on Machine Translation (WMT)}, pages 69--99.

\end{thebibliography}
\bibliographystyle{acl_natbib}

\appendix

\begin{table}[t]
  \centering
  \small
  \begin{tabular}{l@{}c@{}r}
    \toprule
    \multicolumn{1}{l}{Model}                                                  & $\mathit{lr}$              & Time [h] \\
    \midrule
    \multicolumn{1}{l}{MEAT}                                                   & $1\times 10^{-6}$          & 24       \\
    M2M-100 (w/ FT)                                                                                                    \\
    \quad\unidir{En}{De}                                                       & $3\times 10^{-5}$ ($\ast$) & 9        \\
    \quad\unidir{En}{Zh}                                                       & $3\times 10^{-5}$ ($\ast$) & 10       \\
    \quad\unidir{Ro}{En}                                                       & $3\times 10^{-5}$ ($\ast$) & 2        \\
    \quad\unidir{Et}{En}                                                       & $3\times 10^{-5}$ ($\ast$) & 2        \\
    \quad\unidir{Ne}{En}                                                       & $3\times 10^{-5}$ ($\ast$) & 2        \\
    \quad\unidir{Si}{En}                                                       & $3\times 10^{-5}$ ($\ast$) & 2        \\
    \midrule
    \multicolumn{1}{l}{XLM-R$_{\mathrm{Base}}$+MLP (split)}                    & $1\times 10^{-5}$          & 556      \\
    \multicolumn{1}{l}{XLM-R$_{\mathrm{Large}}$+MLP (split)}                   & $1\times 10^{-5}$          & 962      \\
    \multicolumn{1}{l}{\textsc{InfoXLM}$_{\mathrm{Base}}$+MLP (split) \#TD=6}  & $1\times 10^{-5}$          & 273      \\
    \multicolumn{1}{l}{LaBSE+MLP (split)}                                      & $1\times 10^{-5}$          & 443      \\
    \midrule
    \multicolumn{1}{l}{XLM-R$_{\mathrm{Base}}$+MLP (concat)}                   & $1\times 10^{-5}$          & 549      \\
    \multicolumn{1}{l}{XLM-R$_{\mathrm{Large}}$+MLP (concat)}                  & $1\times 10^{-5}$          & 1105     \\
    \multicolumn{1}{l}{\textsc{InfoXLM}$_{\mathrm{Base}}$+MLP (concat) \#TD=6} & $1\times 10^{-5}$          & 386      \\
    \multicolumn{1}{l}{LaBSE+MLP (concat) \#TD=6}                              & $1\times 10^{-5}$          & 272      \\
    \midrule
    \textsc{InfoXLM}$_{\mathrm{Base}}$+MLP (concat) \#TD=1                                                             \\
    \quad\unidir{En}{De}                                                       & $5\times 10^{-5}$          & 69       \\
    \quad\unidir{En}{Zh}                                                       & $5\times 10^{-5}$          & 70       \\
    \quad\unidir{Ro}{En}                                                       & $1\times 10^{-5}$          & 61       \\
    \quad\unidir{Et}{En}                                                       & $5\times 10^{-5}$          & 20       \\
    \quad\unidir{Ne}{En}                                                       & $1\times 10^{-5}$          & 49       \\
    \quad\unidir{Si}{En}                                                       & $1\times 10^{-6}$          & 12       \\
    \midrule
    LaBSE+MLP (concat) \#TD=1                                                                                          \\
    \quad\unidir{En}{De}                                                       & $1\times 10^{-5}$          & 80       \\
    \quad\unidir{En}{Zh}                                                       & $5\times 10^{-5}$          & 65       \\
    \quad\unidir{Ro}{En}                                                       & $1\times 10^{-5}$          & 46       \\
    \quad\unidir{Et}{En}                                                       & $1\times 10^{-5}$          & 22       \\
    \quad\unidir{Ne}{En}                                                       & $5\times 10^{-5}$          & 30       \\
    \quad\unidir{Si}{En}                                                       & $5\times 10^{-6}$          & 15       \\
    \bottomrule
  \end{tabular}
  \caption{
    Best learning rate ($\mathit{lr}$) and the total computation time including learning rate search: ($\ast$) indicates that this parameter is not searched.
  }
  \label{table:computation-time}

  \bigskip

  \centering
  \small
  \begin{tabular}{lr}
    \toprule
    Translation direction & Time [h] \\
    \midrule
    \quad\unidir{En}{De}  & 963      \\
    \quad\unidir{En}{Zh}  & 856      \\
    \quad\unidir{Ro}{En}  & 173      \\
    \quad\unidir{Et}{En}  & 31       \\
    \quad\unidir{Ne}{En}  & 6        \\
    \quad\unidir{Si}{En}  & 7        \\
    \bottomrule
  \end{tabular}
  \caption{Time spent for decoding parallel data for generating synthetic TQE data.}
  \label{table:decoding-time}
\end{table}

\section{Computation Time and Learning Rate}
\label{app:rate}

For the entire experiment, a single V100 GPU was used.
See \Tab{table:computation-time} for the learning rate and training time.
We also show the time spent for decoding source-side of the bilingual parallel corpora for generating synthetic TQE data in \Tab{table:decoding-time}.

The best learning rate was determined among four candidate values.  However, we performed just one training run for each learning rate value.
When fine-tuning M2M-100 for each translation direction, we used a fixed learning rate value and performed it only once.

\end{document}